\pdfoutput=1

\documentclass[11pt]{article}

\usepackage[preprint]{acl}

\usepackage{times}
\usepackage{latexsym}

\usepackage[T1]{fontenc}

\usepackage[utf8]{inputenc}

\usepackage{microtype}

\usepackage{inconsolata}

\usepackage{graphicx}
\usepackage{amsmath}
\usepackage{booktabs}
\usepackage{mathrsfs}
\usepackage{amsfonts}
\usepackage{subcaption}
\usepackage{xspace}
\usepackage{algorithm}
\usepackage{algpseudocode}
\usepackage{enumitem}
\usepackage{pifont}
\usepackage{multirow}
\usepackage{colortbl} 

\newcommand{\md}{Imit-MNMT\xspace}

\title{Extending Multilingual Machine Translation through Imitation Learning}

\author{Wen Lai$^{1,2}$, Viktor Hangya$^{1,2}$, Yingli Shen$^3$, Alexander Fraser$^{1,2}$ \\
        $^1$ Center for Information and Language Processing, LMU Munich, Germany \\ 
        $^2$ Munich Center for Machine Learning, Germany \\
        $^3$ Minzu University of China, China \\
        \tt lavine@cis.lmu.de}

\begin{document}
\maketitle
\begin{abstract}
Despite the growing variety of languages supported by existing multilingual neural machine translation (MNMT) models, most of the world's languages are still being left behind.
We aim to extend large-scale MNMT models to incorporate a new language, enabling translations between this new language and all previously supported languages, even in the challenging scenario where only a parallel corpus between the new language and English is available.
Previous methods, such as continued training on parallel data including the new language, often suffer from catastrophic forgetting, which degrades performance on other languages.
We propose a novel approach \md which treats this task as an imitation learning problem, a technique widely used in computer vision but less explored in natural language processing.
Specifically, we leverage an expert model to generate pseudo-parallel corpora between the new language and the existing languages.
We then introduce a data distribution imitation strategy using language-specific weighting, alongside a translation behavior imitation mechanism.
Extensive experiments show that our approach significantly improves translation performance between the new and existing languages while mitigating catastrophic forgetting.
\end{abstract}

\section{Introduction}
\label{sec:intro}
Recent advancements in multilingual neural machine translation (MNMT) have made significant strides in supporting a wide range of languages within a single model.
For example, the \textit{m2m\_100} model~\citep{fan2021beyond} enables translation between 100 languages, while the \textit{nllb} model~\citep{costa2022no} extends this to over 200 languages.
However, with approximately 7,000 languages spoken worldwide\footnote{\url{https://www.ethnologue.com}}, the vast majority of language pairs lack the resources required for training machine translation models.
Extending MNMT models to accommodate more languages remains a significant challenge, increasingly drawing research attention~\cite{sun-etal-2023-efficiently}.

Existing methods for extending MNMT models to new languages can be broadly categorized into three approaches:
\textbf{(i)} continued training by integrating the original MNMT model's data with data from the new language~\cite{gu-etal-2022-continual, sun-etal-2023-efficiently, imanigooghari-etal-2023-glot500, lai-etal-2024-llms};
\textbf{(ii)} extending the model's vocabulary~\cite{imamura2022extending, imanigooghari-etal-2023-glot500} or introducing small layers into the existing MNMT model~\cite{chronopoulou-etal-2023-language, pires-etal-2023-learning, khassanov2024dual}; and
\textbf{(iii)} extending the MNMT model to support translation between the new language and one of the original languages~\cite{zhao2022life, liu-etal-2023-continual, huang-etal-2023-knowledge}.
While promising, (i) typically requires access to the original MNMT training data, and (ii) increases model parameters with each additional language, limiting scalability. (iii), meanwhile, often focuses on translation between the new language and a single target language, neglecting performance across the full set of languages supported by the original MNMT model.
Additionally, these methods do not effectively associate the new language with the original languages supported by the MNMT model, which can easily lead to copy problems~\cite{liu-etal-2021-copying} and off-target issues~\cite{zhang-etal-2020-improving}.

Motivated by these challenges, we explore the following research questions: \\
\textbf{Q1}: Can we extend existing MNMT models to a new language in a challenging scenario where only parallel data between the new language and English is available, without access to the original MNMT training data? \\
\textbf{Q2}: If so, can the extended MNMT model improve translation performance for the new language pairs while preserving performance on the original language pairs?

In this paper, we reformulate the task of extending MNMT models to new languages as an imitation learning problem.
Imitation learning (i.e., learning from demonstrations;~\citealp{hussein2017imitation}) aims to replicate expert behavior based on demonstrations and has shown efficacy in various domains~\cite{gavenski2024imitation,zare2024survey}.
However, its application in NLP tasks has been limited~\cite{wei-etal-2019-imitation,lin-etal-2020-autoregressive,yao-etal-2020-imitation,shi-etal-2022-text}.
Given an expert MNMT model, we randomly select $k$ languages already supported by the expert.
Using the expert model and beam search~\cite{post-vilar-2018-fast}, we generate pseudo-parallel data between English and these $k$ languages.
The learner model is then trained to mimic the data distribution using a language weighting strategy and to imitate the translation behavior of the expert model.

We conduct a comprehensive evaluation to validate the effectiveness of \md by extending the \textit{m2m\_100} model~\cite{fan2021beyond} to four new languages with distinct scripts (Arabic, Cyrillic, Devanagari, and Latin) on Flores-200 benchmark~\cite{costa2022no}.
The FLORES-200 benchmark is well-suited for this task due to its broad coverage of low-resource languages, including the four new languages. Additionally, it includes all the languages supported by \textit{m2m\_100}, which facilitates the evaluation of performance between the new language and the original languages.
Experimental results demonstrate that our method not only improves translation performance for the new languages across all 100 languages supported by the \textit{m2m\_100} model but also maintains performance on the original language pairs.

In summary, our contributions are as follows:
(i) We propose a novel framework (\md) that extends large-scale MNMT models to new languages using imitation learning, relying only on parallel data between English and the new language. To the best of our knowledge, this is the first work to apply imitation learning to this task.
(ii) Our method mitigates the copy problem and off-target translation issues, which are common challenges in large-scale multilingual models.
(iii) We observe that MNMT models exhibit a degree of script-based transfer, whereby extending a new language also enhances the translation performance of other languages that share the same script.
\section{Background and Related Work}
\label{sec:related}

\subsection{Multilingual Machine Translation}
\label{background:task_definition}

Multilingual Neural Machine Translation (MNMT) refers to a many-to-many mapping function that enables translation across multiple languages using a single model~\citep{johnson-etal-2017-googles, fan2021beyond, costa2022no, bala2024multilingual}.
Formally, given an MNMT dataset with $N$ translation pairs $\mathbb{D}=\{(\boldsymbol{x}_i,\boldsymbol{y}_i), i \in 1 \cdots N\}$, the training loss is defined as:
\vspace{-1em}
\begin{equation}
    \label{eq:mnmt_loss}
    \mathcal{L}_{\mathit{MNMT}}=-\sum_{\boldsymbol{x},\boldsymbol{y} \in \mathbb{D}}\sum_{j=1}^{J} \log\ p_{\theta}(y_j|\boldsymbol{y}_{<j},\boldsymbol{x})
\vspace{-0.7em}
\end{equation}
where $\boldsymbol{x} = x_1, x_2, \cdots, x_I$ represents a source sentence of length $I$, and $\boldsymbol{y} = y_1, y_2, \cdots, y_J$ is the corresponding target sentence of length $J$.

With $L$ languages, an MNMT model can translate between $L \times (L-1)$ language pairs.
However, when extending the model to include a new language $\mathcal{L}_{new}$, the number of supported language pairs increases to $(L+1) \times L$. 
Traditional methods for extending MNMT models typically involve vocabulary expansion~\citep{ko-etal-2021-adapting, nguyen2023seallms, imanigooghari-etal-2023-glot500} or adding new layers to the model~\cite{chronopoulou-etal-2023-language, pires-etal-2023-learning, khassanov2024dual}.
While these methods partially address this challenge, they often require access to the training data of the original MNMT model or suffer from severe catastrophic forgetting.
In this paper, we aim to extend the model to a new language using only parallel corpora between the new language and English, without causing 
catastrophic forgetting.

\subsection{Imitation Learning}
Imitation Learning (IL) is a framework in which a learner model acquires skills by mimicking the behavior of an expert model~\cite{gavenski2024imitation}.
The learner is provided with expert demonstrations, which it uses to learn an optimal policy for decision-making tasks.
IL has been widely applied in fields such as robotics~\cite{kawaharazuka2024robotic}, autonomous driving~\cite{cheng2024pluto}, and vision-based tasks~\cite{guo2024lasil, wang2024genh2r}.
However, its application in natural language processing (NLP), particularly in machine translation, is relatively underexplored~\cite{lin-etal-2020-autoregressive, yao-etal-2020-imitation, shi-etal-2022-text}.

The goal in IL is to find a policy $\pi$ that maps states $s \in S$ to actions $a \in A$, such that the learner’s actions replicate those of the expert.
Given a set of expert demonstrations $\{(s_i, a_i)\}_{i=1}^{N}$, where $s_i$ are states (input sentences) and $a_i$ are the corresponding actions (translated sentences), the objective is to minimize the discrepancy between the learner’s and the expert’s actions.
This is typically done by minimizing a loss function $\mathcal{L}(\pi(s), a)$ (e.g., cross-entropy loss).
In this paper, we apply imitation learning to the MNMT extension task by treating the original MNMT model as the expert and the extended model as the learner.
\section{Methods}
\label{sec:methods}

\begin{figure*}[!ht]
\centering
        \includegraphics[width=\linewidth]{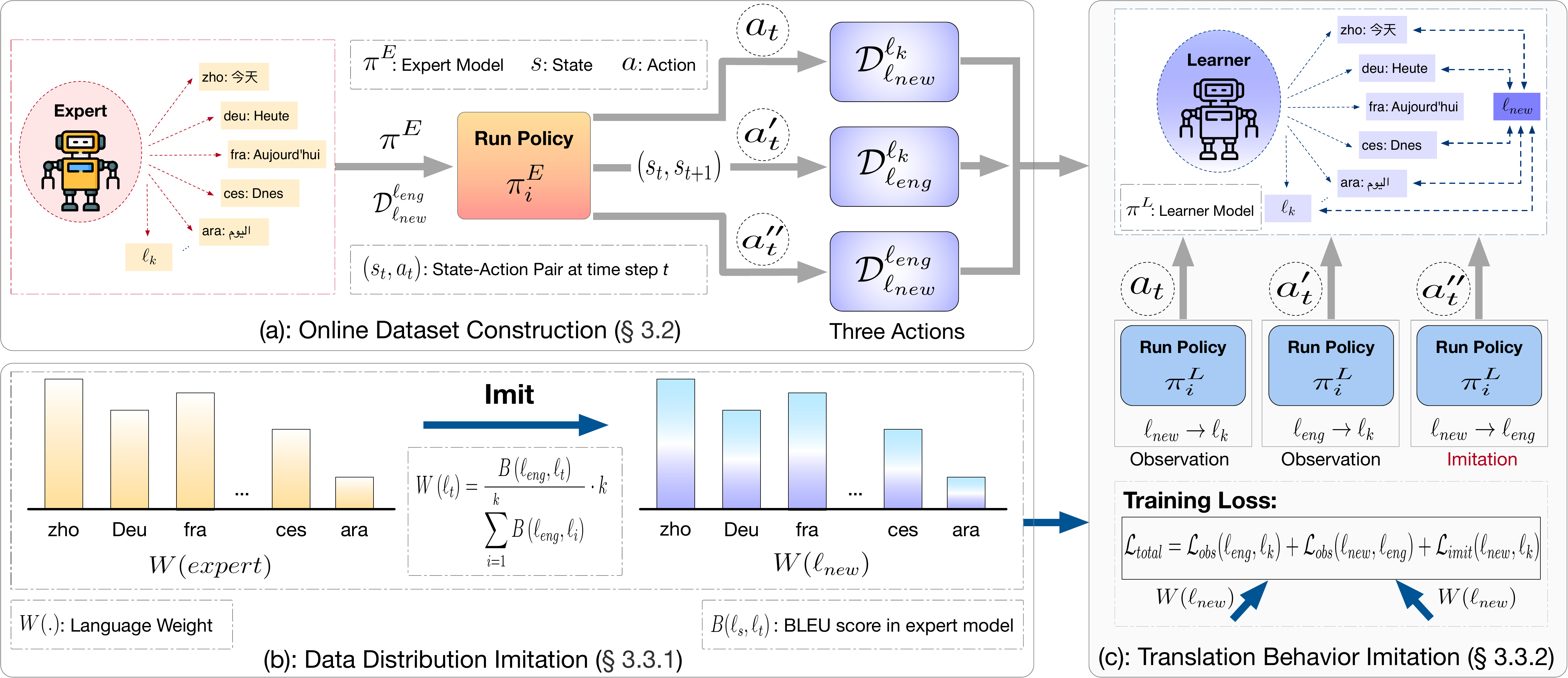}
	\caption{\md consists three parts. In each time step $t$, the expert emit three actions corresponds to parallel data between $(\ell_{new}, \ell_{eng})$, $(\ell_{eng}, \ell_{k})$ and $(\ell_{new}, \ell_{k})$. Then, the data distribution for $\ell_{new}$ in the learner model is imitated by using a language weighting strategy. Finally, the translation behavior is imitated by two types of loss function ($\mathcal{L}_{obs}$ and $\mathcal{L}_{imit}$), where $\mathcal{L}_{obs}$ is designed for mitigating catastrophic forgetting.
 }
    \label{fig:imit_MNMT}
 \vspace{-1em}
\end{figure*}

The goal of our method is to extend a pretrained MNMT model to support translations between an additional new language $\ell_{new}$ and the existing $L$ languages, without sacrificing performance on the already supported $L \times (L-1)$ language pairs by using only parallel data between the new language and English.
We first introduce the overview of taking extending MNMT task as an imitation learning problem (Section~\ref{sec:overview_imit}).
Then, we present a detailed description of our imitation learning framework, including online dataset construction (Section~\ref{sec:data_construct}) and imitation learning process (Section~\ref{sec:imit}), which contains the imitation of data distribution (Section~\ref{sec:data_dis}) and translation behavior (Section~\ref{sec:trans_imit}).
Figure~\ref{fig:imit_MNMT} illustrates the framework of \md.

\subsection{Problem Definition}
\label{sec:overview_imit}
We reformulate the problem within the imitation learning framework, treating the original MNMT model as the ``expert'' that the extended model (the ``learner'') should imitate when translating between the new and existing languages. This helps maintain the existing translation knowledge while extending the model to new tasks.
The states, actions and policy can be defined as:
\begin{itemize}[topsep=0pt, partopsep=0pt]
    \setlength{\itemsep}{-5pt}
    \item \textbf{States ($s$):} A state corresponds to a source sentence, either in one of the original languages $\ell_1, \ell_2, ..., \ell_L$ or in the new language $\ell_{new}$. These sentences are input sequences for the translation system (i.e., for the expert and/or the learner).
    \item \textbf{Actions ($a$):} Actions represent the translation of a source sentence into a target language. The expert model produces actions denoted as $a^E$, while the learner model produces actions $a^L$. The learner's goal is to produce translations similar to those of the expert.
    \item \textbf{Policy ($\pi$):} The policy refers to the translation model itself, which generates translations based on the state-action $(s_t,a_t)$ pairs in time step $t$. The expert policy $\pi^{E}$ corresponds to the original MNMT model, and the learner policy $\pi^{L}$ corresponds to the extended MNMT model.
\end{itemize}

The imitation learning objective is to minimize the divergence between the learner’s actions (translations) and those of the expert model, for both existing language pairs and the new language pairs involving $\ell_{new}$. Specifically, we optimize the cross-entropy loss between the expert’s and learner’s translations:
\vspace{-1em}
\begin{equation}
\label{eq:imit_overall}
\mathcal{L}(\pi^{L})=\mathbb{E}_{\boldsymbol{y} \mid \boldsymbol{x} \sim \mathcal{D}}\left[\sum_{t=1}^{T} \ell^{\pi^{*}}\left(\boldsymbol{y}_{<t}, \boldsymbol{x} ; \pi^{E}\right)\right]
\vspace{-0.7em}
\end{equation}
where $\ell^{\pi^*}(\cdot; \pi)$ is the next-token loss function measuring the discrepancy between the expert model ($\pi^{E}$) and learner model ($\pi^{L}$) given the prior context $\{\boldsymbol{y_{<t}}, \boldsymbol{x}\}$.  

\subsection{Online Dataset Construction}
\label{sec:data_construct}
English, as the most resourceful language in the world, often has easily accessible parallel corpora with other languages.
Therefore, our scenario is to extend the MNMT model using only the parallel corpus between the new language and English.
As a foundation for imitation learning, we first construct multi-parallel data between the new language and the original languages in an online mode.

Given a parallel corpus $\mathcal{D}_{\ell_{new}}^{\ell_{eng}}$ between a new language $\ell_{new}$ and English $\ell_{eng}$, we randomly select $k$ languages ($\mathscr{L}_{k-lang}$) already supported by the expert MNMT model to construct a pseudo $k$-way parallel dataset $\mathcal{\hat{D}}_{\ell_{new}}^{\mathscr{L}_{k-lang}} = \{\mathcal{\hat{D}}_{\ell_{new}}^{\ell_k} : k \in \mathscr{L}_{k-lang}\}$ between the new language and the $k$ languages, utilizing beam search from the MNMT model.
In addition, we also construct a pseudo $k$-way parallel dataset $\mathcal{\hat{D}}_{\ell_{eng}}^{\mathscr{L}_{k-lang}} = \{\mathcal{\hat{D}}_{\ell_{eng}}^{\ell_k} : k \in \mathscr{L}_{k-lang}\}$ between English and the $k$ languages.
The construction process of $\mathcal{\hat{D}}_{\ell_{new}}^{\ell_k}$ can be formulated as:
\vspace{-0.7em}
\begin{equation}
\label{eq:cons_pseudo}
\bigcup_{\boldsymbol{x}^{\ell_{new}} \mid \boldsymbol{x}^{\ell_{eng}} \in \mathcal{D}_{\ell_{new}}^{\ell_{eng}}}  gen\left(\pi^{E},\boldsymbol{x}^{\ell_{eng}}, \ell_k \right) 
\vspace{-0.7em}
\end{equation}
where $gen(\cdot)$ is the beam search function.
Note that the parameters of $\pi^{E}$ are not updated during the generation process and the $k$ languages are resampled in each batch.

\subsection{Extending MNMT as Imitation Game}
\label{sec:imit}
After constructing the k-way parallel corpora for the new language, we can initiate the imitation process.
More specifically, it is divided into two parts: Imitation of data distribution (DDI; Section~\ref{sec:data_dis}) and translation behavior (TBI; Section~\ref{sec:trans_imit}).

\subsubsection{Data Distribution Imitation}
\label{sec:data_dis}
The expert MNMT model is trained on a set of parallel corpora consisting of multiple language pairs.
However, the size of these corpora is often imbalanced, which leads to suboptimal performance in certain languages~\citep{lai-etal-2023-mitigating}.
To address this issue in the learner model, we reduce the influence of low-performing languages during the imitation learning process.

In general, we assume that the importance of a given language during training aligns closely with the performance of the expert model on that language.
Language pairs that demonstrate exceptional performance in the expert MNMT model are expected to yield high-quality pseudo data when one side (either source or target) is replaced with the new language being integrated into the MNMT model.
Conversely, low-performing language pairs are anticipated to produce lower-quality data. 
To operationalize this, we compute the BLEU score of the expert model for each original language paired with English using the FLORES-101 \texttt{devtest} dataset~\cite{goyal-etal-2022-flores}.
Subsequently, we assign higher weights to languages with superior BLEU scores, thereby approximating their data distribution in the expert model during the training process of the learner.

More specifically, the weight of a non-English language $\ell_t$ can be calculated as:
\vspace{-1em}
\begin{equation}
\label{eq:weight_assign}
W\left(\ell_t\right) = \frac{B\left(\ell_{eng}, \ell_t\right)}{\displaystyle\sum_{i=1}^{k} B\left(\ell_{eng}, \ell_{i}\right)}\cdot k
\vspace{-0.7em}
\end{equation}
where $B\left(\ell_s, \ell_t\right)$ is the BLEU score for the language pair from $\ell_s$ to $\ell_t$.
This weight distribution is then applied in the subsequent training steps.

\subsubsection{Translation Behavior Imitation}
\label{sec:trans_imit}
As described in Eq.~(\ref{eq:imit_overall}), the primary objective of our framework is to minimize the divergence between the expert model $\pi^{E}$ and the learner model $\pi^{L}$.
This objective is quantified by the next-token loss function $\ell^{\pi^*}(\cdot; \pi)$, which measures the discrepancy between the actions (translations) produced by the expert model and the learner model, conditioned on the prior context $\{\boldsymbol{y_{<t}}, \boldsymbol{x}\}$.
To operationalize this framework, we decompose the translation behavior imitation process into two distinct components, each corresponding to different types of data used for training the learner model.

\paragraph{Observation Loss $\mathcal{L}_{obs}$:} The observation loss pertains to the training of the learner model using data from the observation dataset ($\mathcal{D}_{\ell_{new}}^{\ell_{eng}}$ and $\mathcal{\hat{D}}_{\ell_{eng}}^{\ell_{k}}$), which consists of language pairs already supported by the expert model.

\vspace{-1em}
\begin{small}
    \begin{equation}
        \label{eq:loss_obs}
        \mathcal{L}_{obs}\left(\ell_1, \ell_2\right) = \mathbb{E}_{\boldsymbol{y}, \boldsymbol{x} \sim \mathcal{D}_{\ell_1}^{\ell_2}} \left[\sum_{t=1}^{T} \ell^{\pi^*}\left(\boldsymbol{y}_{<t}, \boldsymbol{x}; \pi^{L}\right)\right]
        \vspace{-0.7em}
    \end{equation}
\end{small}

Here, $\mathcal{L}_{obs}$ captures the learner model's performance on the observation dataset, where the loss is computed with respect to the ground-truth sequence from $\mathcal{D}_{\ell_1}^{\ell_2}$, aligning with $\ell^{\pi^*}$.

\paragraph{Imitation Loss $\mathcal{L}_{imit}$:} The imitation loss is applied to the pseudo-parallel dataset ($\mathcal{\hat{D}}_{\ell_{new}}^{\ell_{k}}$) generated by the expert model for the new language. In this case, the learner model's output is compared to the pseudo-reference generated by the expert model, as formulated in:

\vspace{-1em}
\begin{small}
    \begin{equation}
        \label{eq:loss_imit}
        \mathcal{L}_{imit}\left(\ell_1, \ell_2\right) = \mathbb{E}_{\boldsymbol{\hat{y}}, \boldsymbol{x} \sim \mathcal{\hat{D}}_{\ell_1}^{\ell_2}} \left[\sum_{t=1}^{T} \ell^{\pi^*}\left(\boldsymbol{\hat{y}}_{<t}, \boldsymbol{x}; \pi^{L}\right)\right]   
        \vspace{-0.7em}
    \end{equation}
\end{small}

Here, $\mathcal{L}_{imit}$ represents the learner model's ability to imitate the expert's translations (pseudo-targets) for the new language. This imitation aligns with $\ell^{\pi^*}$ in Eq.~(\ref{eq:imit_overall}) but is specifically applied to the pseudo-parallel data $\mathcal{\hat{D}}_{\ell_1}^{\ell_2}$.

\paragraph{Overall Training Objective:} The total training objective $\mathcal{L}_{total}$ combines both the observation loss and the imitation loss, weighted according to the significance of the language pairs based on their BLEU scores:

\vspace{-1.8em}
\begin{small}
    \begin{multline}
    \label{eq:loss_all}
    \mathcal{L}_{total} = \mathcal{L}_{obs}\left(\ell_{eng}, \ell_{new}\right) + \sum_{i=1}^{k} W(\ell_{k}) \cdot \mathcal{L}_{obs}\left(\ell_{eng}, \ell_{k}\right) \\
         + \sum_{i=1}^{k} W(\ell_{k}) \cdot \mathcal{L}_{imit}\left(\ell_{new}, \ell_{k}\right)
\vspace{-1.8em}
\end{multline}
\end{small}

This objective ensures that the learner model $\pi^{L}$ benefits from both the observation data (i.e., mitigating catastrophic forgetting problem) and the pseudo-parallel data for the new language, while retaining the expert model's knowledge on the original tasks.
\begin{table*}[h]
\small
\centering
\begin{tabular}{l|ccc|ccc|ccc|ccc}
\toprule
\multicolumn{13}{c}{\textbf{Translation from new languages to original languages}} \\
\midrule
  & \multicolumn{3}{c}{\textbf{prs\_Arab}} & \multicolumn{3}{c}{\textbf{tat\_Cyrl}} & \multicolumn{3}{c}{\textbf{mag\_Deva}} & \multicolumn{3}{c}{\textbf{lus\_Latn}} \\
\cmidrule{2-13}
  & Low  & Mid & High & Low  & Mid & High & Low  & Mid & High & Low  & Mid & High \\
\hline
m2m\_100      & 1.03 & 2.47 & 3.94  & 0.71 & 1.85 & 2.97  & 0.79 & 1.34 & 2.31 & 2.79 & 3.88 & 4.84  \\
Finetune      & 1.50 & 2.52 & 4.12  & 1.03 & 1.91 & 3.21  & 1.17 & 1.72 & 2.68 & 2.92 & 4.26 & 5.05  \\
Extend\_Vocab & 1.65 & 2.71 & 4.35  & 1.25 & 2.15 & 3.42  & 1.51 & 1.98 & 3.06 & 3.64 & 4.80 & 5.80  \\
KD            & 1.85 & 2.81 & 4.95  & 1.53 & 2.42 & 3.94  & 1.81 & 2.22 & 3.18 & 3.94 & 4.89 & 6.11  \\
Adapter       & 1.92 & 2.92 & 5.15  & 1.62 & 2.61 & 4.14  & 2.12 & 2.71 & 3.63 & 3.65 & 5.48 & 6.50  \\
MMA           & 2.07 & 3.15 & 5.50  & 2.11 & 2.96 & 4.96  & 2.90 & 3.21 & 3.99 & 3.96 & 5.75 & 7.16  \\
\midrule
\md           & \textbf{4.71}$^\ddag$ & \textbf{7.04}$^\ddag$ & \textbf{12.26}$^\ddag$ & \textbf{4.69}$^\ddag$ & \textbf{6.67}$^\ddag$ & \textbf{10.75}$^\ddag$ & \textbf{4.48}$^\ddag$ & \textbf{5.67}$^\ddag$ & \textbf{9.01}$^\ddag$ & \textbf{6.83}$^\ddag$ & \textbf{9.91}$^\ddag$ & \textbf{15.38}$^\ddag$  \\
\bottomrule
\end{tabular}
\vspace{0.5em}
\centering
\begin{tabular}{l|ccc|ccc|ccc|ccc}
\toprule
\multicolumn{13}{c}{\textbf{Translation from original languages to new languages}} \\
\midrule
  & \multicolumn{3}{c}{\textbf{prs\_Arab}} & \multicolumn{3}{c}{\textbf{tat\_Cyrl}} & \multicolumn{3}{c}{\textbf{mag\_Deva}} & \multicolumn{3}{c}{\textbf{lus\_Latn}} \\
\cmidrule{2-13}
  & Low  & Mid & High & Low  & Mid & High & Low  & Mid & High & Low  & Mid & High \\
\hline
m2m\_100      & -        & -       & -        & -        & -       & -        & -        & -       & -        & -        & -        & -       \\
Finetune      & 0.74 & 0.97 & 1.24  & 0.64 & 0.77 & 1.38 & 0.76 & 0.99 & 1.82 & 0.99 & 1.27 & 1.95  \\
Extend\_Vocab & 0.92 & 1.51 & 1.70  & 0.77 & 0.93 & 1.86 & 0.84 & 1.40 & 2.32 & 1.28 & 1.51 & 2.40  \\
KD            & 1.04 & 1.76 & 1.95  & 0.94 & 1.33 & 2.10 & 1.05 & 1.80 & 2.65 & 1.08 & 1.79 & 2.64  \\
Adapter       & 1.28 & 2.95 & 2.06  & 1.22 & 1.30 & 2.35 & 1.19 & 1.92 & 2.60 & 1.25 & 1.99 & 3.10  \\
MMA           & 1.55 & 3.06 & 2.26  & 1.56 & 1.64 & 2.69 & 1.55 & 2.08 & 3.00 & 1.49 & 2.28 & 3.55  \\
\midrule
\md           & \textbf{4.32}$^\ddag$ & \textbf{6.36}$^\ddag$ & \textbf{10.51}$^\ddag$ & \textbf{4.16}$^\ddag$ & \textbf{6.03}$^\ddag$ & \textbf{9.11}$^\ddag$ & \textbf{4.35}$^\ddag$ & \textbf{6.17}$^\ddag$ & \textbf{8.67}$^\ddag$ & \textbf{6.22}$^\ddag$ & \textbf{9.31}$^\ddag$ & \textbf{13.36}$^\ddag$ \\
\bottomrule
\end{tabular}
\vspace{-0.8em}
\caption{\label{tab:q1}
\textbf{Main Results (the answer of Q1)}: Average BLEU scores for different categories in two directions on  FLORES-200 benchmark. The original languages indicates the languages already supported in m2m\_100. 
$^\ddag$ denotes significant over original \textit{m2m\_100} model at 0.05/0.01, evaluated by bootstrap resampling~\cite{koehn-2004-statistical}.
}
\vspace{-1.8em}
\end{table*}
\begin{table*}[h]
\small
\centering
\begin{tabular}{l|ccccccccc}
\toprule
\multicolumn{10}{c}{\textbf{Extended model trained from prs\_Arab to original languages}} \\
\midrule
              & L2L  & L2M  & L2H  & M2L   & M2M   & M2H   & H2L  & H2M   & H2H   \\
\midrule
m2m\_100      & \textbf{1.23} & \textbf{2.37} & \underline{7.43} & \underline{12.29} & \underline{12.25} & \textbf{14.25} & \underline{8.36} & \underline{10.27} & \textbf{15.74} \\
Finetune      & 0.65 & 0.50 & 0.88 & 1.19  & 2.46  & 3.55  & 2.51 & 2.16  & 4.60  \\
Extend\_Vocab & 0.27 & 0.94 & 0.62 & 0.86  & 1.84  & 1.76  & 1.19 & 1.41  & 1.88  \\
KD            & 0.63 & 0.89 & 3.69 & 6.85  & 7.98  & 8.62  & 4.35 & 4.57  & 8.40  \\
\midrule
\md           & \underline{1.15} & \underline{2.20} & \textbf{7.68} & \textbf{12.97} & \textbf{12.66} & \underline{14.06} & \textbf{8.43} & \textbf{10.31} & \underline{15.42} \\
$\Delta$      & -0.08 & -0.17 & +0.25 & +0.68 & +0.41 & -0.19 & +0.07 & +0.04 & -0.32 \\
\midrule\midrule
\multicolumn{10}{c}{\textbf{Extended model trained from original languages to mag\_Deva}} \\
\midrule
              & L2L  & L2M  & L2H  & M2L   & M2M   & M2H   & H2L  & H2M   & H2H   \\
\midrule
m2m\_100      & \textbf{1.03} & \textbf{1.61} & \textbf{5.38} & \underline{10.22} & \underline{10.73} & \textbf{18.32} & \underline{5.19} & \textbf{13.28} & \underline{16.25} \\
Finetune      & 0.31 & 0.54 & 0.75 & 1.40  & 2.83  & 2.37  & 1.41 & 2.88  & 2.70  \\
Extend\_Vocab & 0.32 & 0.52 & 0.35 & 1.32  & 1.73  & 1.68  & 0.90 & 2.09  & 1.58  \\
KD            & 0.62 & 0.92 & 2.03 & 4.15  & 5.02  & 7.70  & 1.94 & 6.37  & 8.81  \\
\midrule
\md           & \underline{0.94} & \underline{1.27} & \underline{5.06} & \textbf{10.41} & \textbf{11.52} & \underline{17.85} & \textbf{5.24} & \underline{13.14} & \textbf{16.34} \\
$\Delta$      & -0.09 & -0.34 & -0.32 & +0.19 & +0.79 & -0.47 & +0.05 & -0.14 & +0.09 \\
\bottomrule
\end{tabular}
\vspace{-0.5em}
\caption{
\label{tab:q2}
\textbf{Main Results (the answer of Q2):} Average BLEU scores of the extended model on 9 original language pairs grouped by available resources on both source and target sizes (\textbf{L}ow, \textbf{M}id, \textbf{H}igh). $\Delta$ indicates the difference between \textit{m2m\_100} and our approach. \textbf{Bold} and \underline{underlined} numbers indicates the best and second-best results respectively. We do not include the results of \textit{Adapter} and \textit{MMA} method, because they fix the parameters of original model and the results are the same as in original m2m\_100.
}
\vspace{-1.5em}
\end{table*}

\section{Experiments}
\label{sec:exp}

\noindent \textbf{Datasets.}
We conduct experiments on four new languages across different scripts\footnote{We follow~\citet{costa2022no} to use three-letter ISO 639-3 code with ISO 15924 script subtags.}:
Dari (\textit{Arabic script}; \texttt{prs\_Arab}), Tatar (\textit{Cyrillic script}; \texttt{tat\_Cyrl}), Magahi (\textit{Devanagari script}; \texttt{mag\_Deva}) and Mizo (\textit{Latin script}; \texttt{lus\_Latin}).
We use parallel corpora between new languages and English from the minded \textit{nllb} dataset\footnote{\url{https://huggingface.co/datasets/allenai/nllb}} as the training data.
We filter out sentences longer than 120 tokens and preprocess all data using sentencepiece~\citep{kudo-richardson-2018-sentencepiece}.
More details of the data can be found in Appendix~\ref{appendix:datasets}.

\textbf{Baselines.}
We compare our method to the following baselines.
(1) \textbf{m2m\_100}: Using the original \textit{m2m\_100} model~\citep{fan2021beyond}.
(2) \textbf{Finetune}: Fine-tuning m2m\_100 model on the parallel data between new language and English.
(3) \textbf{Extend\_Vocab}: Extending the vocabulary of the original m2m\_100 model with tokens of the new language, then continue training using the same data as for \textit{Finetune}~\citep{imanigooghari-etal-2023-glot500}.
(4) \textbf{KD}: Knowledge distillation approach~\citep{castellucci-etal-2021-learning,zhang-etal-2023-towards-understanding} that distill the expert model.
(5) \textbf{Adapter}: Train an additional language-specific layer for the new language~\citep{lai-etal-2022-4,chronopoulou-etal-2023-language}.
(6) \textbf{MMA}: Build a shallow mini-model from a fraction of a large model’s parameters~\cite{marchisio-etal-2023-mini}.
Although many recent methods have been proposed for extending MNMT or multilingual models, they are not suitable as our baselines due to differences in usage scenarios. A detailed explanation of our baseline selection is provided in the Appendix~\ref{appendix:baseline_select_criteria}.

\textbf{Implementation.}
We use the \texttt{m2m\_100 (1.2B)} model\footnote{We do not choose to use the NLLB model~\cite{costa2022no} because it already supports translation for the new language, whereas we need to evaluate the translation between the new language and all of the original 100 languages.} as the basis of the baselines and \md, released in the HuggingFace repository~\citep{wolf-etal-2020-transformers}.
For \textit{Adapter} training, we use the implementation from~\citep{lai-etal-2022-4}.
We implement \textit{Extend\_Vocab} based on~\citet{imanigooghari-etal-2023-glot500}; To ensure a fair comparison, we maintained a consistent extended vocabulary size of 23,288.
We implement \textit{MMA} by ourself since they do not make their code publicly available.
It is worth highlighting that both the \textit{Extend\_Vocab}, \textit{Adapter} and \textit{MMA} baselines introduce additional parameters.
More details of the model configuration can be found in Appendix~\ref{appendix:model_config}.

\textbf{Evaluation.}
We measure case-sensitive detokenized BLEU with SacreBLEU\footnote{\url{https://github.com/mjpost/sacrebleu}}~\citep{post-2018-call}.
Recently, the BLEU score was criticized as an unreliable automatic metric~\citep{kocmi-etal-2021-ship,zerva-etal-2022-disentangling}.
Therefore, we also evaluate our approach using chrF++~\citep{popovic-2017-chrf}.
The corresponding chrF++ scores are shown in Appendix~\ref{appendix:detailed_analysis}.
Similar to~\citet{mohammadshahi-etal-2022-small}, we split the languages based on the amount of available training sentences aligned with English into 3 different categories: Low(L), Mid(M) and High(H).
All results are evaluated on the FLORES-200 benchmark\footnote{\url{https://github.com/facebookresearch/flores/tree/main/flores200}}.
\section{Results}
\label{sec:res}
Tables~\ref{tab:q1} and Table~\ref{tab:q2} present the answers to the two research questions posed in Section~\ref{sec:intro}.
For a more in-depth analysis, please refer to Appendix~\ref{appendix:detailed_analysis}.

\noindent \textbf{Q1: Successful Extension to a New Language}\par
As shown in Table~\ref{tab:q1}, all baseline systems demonstrate improvements over the original m2m\_100 model, though challenges persist, especially for low- and mid-resource languages. We attribute these difficulties to the limited support for such languages in the original m2m\_100 model~\cite{lai-etal-2023-mitigating}. 
Additionally, we observe that translation performance is superior when translating from the new language into the original language, compared to the reverse.
This occurs because the decoder in machine translation models tends to be more sensitive to noisy inputs than the encoder.
Since less data is available for the new language, training an effective decoder from scratch proves challenging.
Furthermore, our approach achieves better results in language pairs where the original language is high-resource, in comparison to pairs involving low- and mid-resource languages.

\noindent \textbf{Q2: Avoiding Catastrophic Forgetting}\par
Table~\ref{tab:q2} demonstrates that all baselines suffer from catastrophic forgetting, where adapting to a new language diminishes performance on original language pairs.
Consequently, the translation quality for these original pairs significantly degrades.
In contrast, our method alleviates this issue, preserving the translation quality for the original language pairs supported by the MNMT model.
This improvement is attributed to our data distribution imitation (Section~\ref{sec:data_dis}), which adjusts translation weights between the new and original languages based on their performance with English.
This approach ensures that previously acquired translation knowledge is not substantially disrupted by the inclusion of the new language.
Our findings indicate that language pairs involving high-resource target languages (e.g., \textit{L2H}, \textit{M2H}, and \textit{H2H}) consistently outperform the other six translation categories.

\noindent \textbf{Statistical Significance Tests}\par
Although the BLEU scores for low-resource languages in Tables~\ref{tab:q1} and~\ref{tab:q2} are relatively low, they remain meaningful.
First, we report the average BLEU score across all low-resource languages.
We observe that a significant portion of the languages supported by the original m2m\_100 model have BLEU scores between 0 and 1 in translation with English. This is the primary reason for the lower BLEU scores in low-resource languages in the extended model (more details are shown in Appendix~\ref{appendix:bleu_dis}). 
Second, we conduct statistical significance tests on low-resource language pairs, finding that the results generated by our method are significantly different from those of the original m2m\_100 model and other baselines.
Notably, the baseline systems produced a substantial number of repetitive and identical words, a problem that our method effectively mitigates (see Section~\ref{sec:copy_and_off} for more details).
\section{Analysis}
\label{sec:analysis}


\subsection{Ablation Study}
\label{sec:ablation}
The training objective function, shown in Eq. (\ref{eq:loss_all}), consists of three loss components. 
We conduct an ablation study on the \texttt{lus\_Latn} extension task to assess the contribution of each loss function. 
Table~\ref{tab:ablation_loss} reveals the following:
(1) The substantial performance difference between \#1 and \#2, with and without $\mathcal{L}_{imit}(\ell_{new},\ell_{k})$, highlights the importance of imitating the translation behavior between the new language and $k$ randomly selected languages for our approach.
(2) While $\mathcal{L}_{obs}(\ell_{new},\ell_{eng})$ and $\mathcal{L}_{obs}(\ell_{eng},\ell_{k})$ in \#1 and \#2 do not significantly alter overall performance, the combination of all three loss functions produces the best results. 
This suggests that preserving the performance of the pivot language, English, is also vital in the imitation learning process.

\begin{table}[!htb]
\small
\centering
\begin{tabular}{c|ccc|ccc}
\toprule
& \textbf{$\mathcal{L}_{1}$} & \textbf{$\mathcal{L}_{2}$} & \textbf{$\mathcal{L}_{3}$} & \textbf{Low} & \textbf{Mid} & \textbf{High}  \\
\midrule
\multirow{3}{*}{\#1} & \ding{51} & \ding{55} & \ding{55} & 0.86 & 0.83 & 1.28  \\
& \ding{55} & \ding{51} & \ding{55} & 0.88 & 0.79 & 1.32  \\
& \ding{55} & \ding{55} & \ding{51} & 0.94 & 1.25 & 2.37  \\
\midrule
\multirow{3}{*}{\#2} & \ding{51} & \ding{51} & \ding{55} & 0.89 & 1.14 & 2.05  \\
& \ding{51} & \ding{55} & \ding{51} & 1.42 & 2.86 & 3.52  \\
& \ding{55} & \ding{51} & \ding{51} & 3.77 & 4.91 & 7.43  \\
\midrule
\#3 (our) & \ding{51} & \ding{51} & \ding{51} & \textbf{6.83} & \textbf{9.91} & \textbf{15.38} \\
\bottomrule
\end{tabular}
\vspace{-0.5em}
\caption{
\label{tab:ablation_loss}
\textbf{Ablation} on loss functions in \texttt{lus\_Latn} extension task. $\mathcal{L}_{1}$, $\mathcal{L}_{2}$ and $\mathcal{L}_{3}$ denotes $\mathcal{L}_{obs}(\ell_{new},\ell_{eng})$, $\mathcal{L}_{obs}(\ell_{eng},\ell_{k})$ and $\mathcal{L}_{imit}(\ell_{new},\ell_{k})$ in Eq. (\ref{eq:loss_all}).
``\ding{51}'' means the loss function is included in the training objective while ``\ding{55}'' means it is not.}
\vspace{-1.8em}
\end{table}

To further evaluate the impact of the dynamic language weighting strategy proposed in Section~\ref{sec:data_dis}, as well as the efficacy of our imitation learning framework (i.e., separating the expert and learner models rather than the \texttt{on-the-fly} mixing approach\footnote{An explanation of the \texttt{on-the-fly} method can be found in the Appendix~\ref{appendix:on_the_fly}.}), we perform an ablation analysis. The results are presented in Table~\ref{tab:ablation_lw_on}. 
Comparisons between \#2 and \#3, and \#4 and \#5, demonstrate that the language weighting (\texttt{LW}) is advantageous in both \md and \texttt{On-the-Fly} scenarios, with particularly strong performance in \md.
Moreover, the comparison between \#3 and \#5 emphasizes the strength of our method, which involves separating the expert model from the learner model and updating the weights independently.

\begin{table}[!htp]
    \small
    \centering
    \begin{tabular}{ll|ccc}
        \toprule
        \multicolumn{5}{c}{\textbf{lus\_Latin $\rightarrow$ original languages}} \\
        \midrule
            &                    & Low  & Mid  & High  \\
        \midrule
        \#1 & m2m\_100           & 2.79 & 3.88 & 4.84  \\
        \midrule
        \#2 & On-the-Fly         & 3.24 & 4.47 & 5.36  \\
        \midrule
        \#3 & On-the-Fly with LW & 3.77 & 4.83 & 5.92  \\
        \midrule
        \#4 & \md w/o LW          & 4.53 & 5.47 & 6.73  \\
        \midrule
        \#5 & \md                 & \textbf{6.83} & \textbf{9.91} & \textbf{15.38} \\
        \bottomrule
    \end{tabular}
    \vspace{-0.5em}
    \caption{
    \label{tab:ablation_lw_on}
    \textbf{Ablation} on the importance of data distribution imitation (ie., language weighting; LW) and framework (i.e, seperate the expert and learner).
    }
    \vspace{-1.8em}
\end{table}

\subsection{Copy and Off-Target Problems}
\label{sec:copy_and_off}
To further demonstrate the effectiveness of our approach, we examine two common challenges in large-scale multilingual models.
The copying problem~\citep{liu-etal-2021-copying} refers to the excessive copying of words from the source language to the target language, rather than accurate translation.
The off-target problem~\citep{zhang-etal-2020-improving} occurs when the MNMT model translates text into an incorrect language.
We evaluate the copy ratio (CR) and off-target ratio (OTR) as introduced by~\cite{lai-etal-2023-mitigating}, with further details provided in Appendix~\ref{appendix:cr_otr}.

\begin{table}[!htb]
\small
\centering
\resizebox{\columnwidth}{!}{
\begin{tabular}{l|ccc||ccc}
\toprule
& \multicolumn{3}{c}{\textbf{CR} $\downarrow$} & \multicolumn{3}{c}{\textbf{OTR} $\downarrow$} \\
\cmidrule(lr){2-4} \cmidrule(lr){5-7}
              & Low  & Mid  & High & Low  & Mid  & High \\
\midrule
m2m\_100      & 0.32 & 0.28 & 0.24 & 0.41 & 0.36 & 0.32 \\
Finetune      & 0.32 & 0.33 & 0.30 & 0.38 & 0.34 & 0.28 \\
Extend\_Vocab & 0.30 & 0.32 & 0.27 & 0.38 & 0.31 & 0.27 \\
KD            & 0.28 & 0.29 & 0.24 & 0.34 & 0.26 & 0.26 \\
Adapter       & 0.27 & \underline{0.24} & 0.24 & \underline{0.29} & 0.22 & 0.23 \\
MMA           & \underline{0.25} & 0.26 & \underline{0.22} & 0.32 & \underline{0.20} & \underline{0.20} \\
\midrule
Our           & \textbf{0.15} & \textbf{0.11} & \textbf{0.08} & \textbf{0.16} & \textbf{0.09} & \textbf{0.06} \\
\bottomrule
\end{tabular}
}
\caption{
\label{tab:copy_off}
\textbf{Copy and Off-Target Problem}: results of copy ratio (CR) and off-target ratio (OTR). A lower value indicates better performance of the model. 
\textbf{Bold} and \underline{underlined} numbers indicates the best and second-best results respectively.
}
\vspace{-1.2em}
\end{table}
Table~\ref{tab:copy_off} presents the CR of the extended model for translations from \texttt{prs\_Arab} to the original languages, and the OTR for translations from the original languages to \texttt{tat\_Cyrl}.
We highlight the following findings:
(i) Our approach effectively reduces both \textit{CR} and \textit{OTR}, addressing these challenges.
(ii) The results align with the comparisons presented in Tables~\ref{tab:q1} and~\ref{tab:q2}, demonstrating that our method outperforms other baselines in extending to new languages.
This indicates that our model effectively integrates the new language’s representation, leading to reductions in both \textit{CR} and \textit{OTR}.

\subsection{Language Script Analysis}
\label{sec:lang_script_anals}
Table~\ref{tab:lang_script} presents the average BLEU scores of the extended model across different scripts for new languages.
We observe a significant improvement in translation performance for the script associated with the new language, while performance for other scripts shows a slight decline compared to the original m2m\_100 model.
This is likely because, under the imitation learning framework, our approach focuses more on the new language's script, where the parallel data between the new language and English is real, while other scripts are represented by synthetic data.
This observation is consistent with findings from back-translation literature~\cite{sennrich-etal-2016-improving}, which suggest that using real target language data improves model performance, while synthetic data may have a negative effect.

\begin{table}[htp]
\small
\centering
\begin{tabular}{l|cccc}
\toprule
          & \textbf{Arab} & \textbf{Cyrl} & \textbf{Deva} & \textbf{Latn} \\
\midrule
m2m       & 4.62 & 3.25 & 2.61 & 8.49  \\
\midrule
prs\_Arab & \cellcolor{lightgray} 5.36 & 2.93 & 2.45 & 8.34  \\
tat\_Cyrl & 4.45 & \cellcolor{lightgray} 3.88 & 2.27 & 8.29  \\
mag\_Deva & 4.20 & 3.14 & \cellcolor{lightgray} 3.21 & 8.05  \\
lus\_Latn & 4.52 & 3.10 & 2.37 & \cellcolor{lightgray} 10.62 \\
\bottomrule
\end{tabular}
\caption{
\label{tab:lang_script}
Average BLEU scores for extended model translation from new languages to original across different language scripts.
}
\vspace{-1.5em}
\end{table}

\subsection{Further Analysis}
\label{sec:further_ana}
To better understand the proposed approach, we present the BLEU score distribution characteristics (Appendix~\ref{appendix:bleu_dis}) of the extended model and its performance in zero-shot domain transfer (Appendix~\ref{appendix:domain}).
Additionally, we explore several factors that may influence the performance of our method, including model size (Appendix~\ref{appendix:model_size}) and the value of $k$ (Appendix~\ref{appendix:k_value}).
Our key findings are as follows:
(1) \md effectively resolves the issue where BLEU scores for most language pairs in the original m2m\_100 model are concentrated in the $0-1$ range.
(2) \md demonstrates robustness in domain transfer compared to other baseline methods.
(3) The 1.2B model outperforms the 418M model in terms of performance.
(4) A larger $k$ value leads to better performance, though it also increases training time.
Please refer to the corresponding Appendix for more details.
\section{Conclusion}
\label{sec:conclusion}
We introduce \textbf{\md}, an innovative approach that extends MNMT model to new languages without compromising the translation performance of existing language pairs.
Specifically, we present a novel perspective on extending an MNMT model by framing it as an imitation learning process.
Remarkably, our approach leverages only a parallel corpus between the new language and English, without requiring access to the training corpora used during the pre-trained model's training phase.
Our method outperforms several robust baseline systems, demonstrating superior performance.
Additionally, it exhibits language script transfer capabilities and provides notable advantages in addressing the copy and off-target problems.
\section{Limitations}
This work has two limitations.
i) We conducted evaluations solely on the m2m\_100 model. However, our approach is expected to be applicable to other MNMT models or even large language models such as mt5, mbart, LLaMA~\cite{dubey2024llama} etc., and can be extended to various other NLP tasks, including question answering and text generation.
ii) We specifically focused on the scenario of utilizing a parallel corpus only from the new language to English.
However, it is worth noting that there might exist parallel sentence pairs between the new language and other languages as well.
We believe that incorporating additional corpora from other languages has the potential to further enhance the overall performance.

\bibliography{acl,custom}

\appendix
\clearpage
\section{Datasets}
\label{appendix:datasets}
All corpora used in our experiments are publicly available through the NLLB corpus, reconstructed based on mining metadata provided by AllenAI.
We applied the following preprocessing steps:
i) elimination of duplicate sentences;
ii) exclusion of sentences longer than 120 tokens;
iii) removal of sentences identified as incorrect language using a language identification model\footnote{We use the language identification model from \url{https://huggingface.co/facebook/fasttext-language-identification}}.
Table~\ref{tab:app_data_sta} presents corpus statistics before and after preprocessing.
To better understanding of our approach, we conduct additional experiments with four supplementary languages.
Our approach is evaluated using the FLORES-200 dataset.

\begin{table}[!ht]
\small
\centering
\resizebox{\columnwidth}{!}{
\begin{tabular}{lll|rr}
\toprule
&\textbf{Language} & \textbf{Script}     & \textbf{\#Size\_b}  & \textbf{\#Size\_a} \\
\midrule
\multirow{4}{*}{\textbf{\#Main}}& Dari (prs\_Arab)    & Arabic     & 6,294,181  & 4,628,319 \\
& Tatar (tat\_Cyrl)    & Cyrillic   & 10,495,969 & 8,210,329 \\
& Magahi (mag\_Deva)   & Devanagari & 2,221,386  & 1,037,829 \\
& Mizo (lus\_Latn)     & Latin      & 6,979,898  & 5,293,361 \\
\midrule
\multirow{4}{*}{\textbf{\#Compare}}& Dyula (Dyu\_Latn)    & Latin      & 286,391    & 215,084   \\
& Balinese (ban\_Latn) & Latin      & 324,936    & 228,206   \\
& Bemba (bem\_Latn)    & Latin      & 427,159    & 319,595   \\
& Banjar (bjn\_Latn)   & Latin      & 766,894    & 620,376   \\
\bottomrule
\end{tabular}
}
\caption{\label{tab:app_data_sta}
Data statistics (number of sentences) of parallel data between new languages and English.
``\#Main'' refers to the languages used in the main experiments presented in the paper, while ``\#Compare'' represents the languages used in supplementary experiments provided in the \textbf{appendix}. ``\#size\_b'' indicates the size of the corpus before data cleaning, and ``\#size\_a'' refers to the size of the corpus after data preprocessing.
}
\end{table}

\section{Selection Criteria for Baseline}
\label{appendix:baseline_select_criteria}
The baseline selection criterion is that the methods must perform the same task with the same data setup. 
From a data perspective, our method only uses a parallel corpus between the new language and English.
From a task perspective, our objective is to extend an MNMT model to support translations between the new language and all languages initially covered by the model. 
Although many related works extend MNMT models, they do not apply to our case, as discussed in Section~\ref{sec:related}.
For example, \citet{lakew-etal-2018-transfer}, \citet{gu-etal-2022-continual}, and \citet{sun-etal-2023-efficiently} require access to original training or validation data, while \citet{zhao2022life}, \citet{liu-etal-2023-continual}, and \citet{huang-etal-2023-knowledge} focus on specific language pairs, neglecting translation for others.

\section{Model Configuration}
\label{appendix:model_config}
The training process consists of two steps.
First, for each batch, we generate a pseudo-parallel corpus in an online mode using an expert model between the new language and $k$ selected languages. 
Next, we apply imitation learning to replicate translations between the new language and the original set of languages.
To maintain consistency with baseline systems, we set the batch size to 16, the learning rate to 5e-5, and the dropout rate to 0.1.
To mitigate overfitting, we limit imitation learning to a single epoch.
The model was trained on one machine equipped with 4 A100 40GB GPUs.

\section{Detailed Analysis}
\label{appendix:detailed_analysis}
This section provides a detailed analysis of the experiments shown in Tables~\ref{tab:q1} and~\ref{tab:q2}.
In addition to BLEU scores, we employ chrF++~\citep{popovic-2017-chrf} as an evaluation metric.
The results in Tables~\ref{tab:q1_chrf} and \ref{tab:q2_chrf} correspond to Tables~\ref{tab:q1} and \ref{tab:q2}, respectively\footnote{We demonstrate that \md consistently outperforms baseline systems in terms of chrF++, aligning with the BLEU score findings.}.

\begin{table*}[!h]
\small
\centering
\begin{tabular}{l|ccc|ccc|ccc|ccc}
\toprule
\multicolumn{13}{c}{\textbf{Translation from new languages to original languages}} \\
\midrule
  & \multicolumn{3}{c}{\textbf{prs\_Arab}} & \multicolumn{3}{c}{\textbf{tat\_Cyrl}} & \multicolumn{3}{c}{\textbf{mag\_Deva}} & \multicolumn{3}{c}{\textbf{lus\_Latn}} \\
\cmidrule{2-13}
  & Low  & Mid & High & Low  & Mid & High & Low  & Mid & High & Low  & Mid & High \\
\hline
m2m\_100      & 16.75 & 23.00 & 26.31 & 14.17 & 20.93 & 24.31 & 14.89 & 18.64 & 22.50 & 23.85 & 26.20 & 27.76  \\
Finetune      & 19.43 & 23.13 & 26.63 & 16.76 & 21.16 & 24.85 & 17.70 & 20.41 & 23.56 & 24.19 & 26.86 & 28.08 \\
Extend\_Vocab & 20.10 & 23.65 & 27.02 & 18.16 & 21.99 & 25.30 & 19.51 & 21.41 & 24.52 & 25.75 & 27.71 & 29.05 \\
KD            & 20.94 & 23.90 & 27.93 & 19.60 & 22.86 & 26.31 & 20.76 & 22.22 & 24.78 & 26.31 & 27.84 & 29.43 \\
Adapter       & 21.21 & 24.18 & 28.21 & 20.01 & 23.38 & 26.66 & 21.92 & 23.64 & 25.73 & 25.76 & 28.65 & 29.87 \\
MMA           & 21.72 & 24.71 & 28.68 & 21.88 & 24.29 & 27.95 & 24.13 & 24.84 & 26.40 & 26.35 & 29.00 & 30.55 \\
\midrule
\md           & \textbf{27.58} & \textbf{30.43} & \textbf{34.38} & \textbf{27.55} & \textbf{30.05} & \textbf{33.45} & \textbf{27.22} & \textbf{28.90} & \textbf{32.19} & \textbf{30.22} & \textbf{32.87} & \textbf{35.99} \\
\bottomrule
\end{tabular}
\vspace{0.5em}
\centering
\begin{tabular}{l|ccc|ccc|ccc|ccc}
\toprule
\multicolumn{13}{c}{\textbf{Translation from original languages to new languages}} \\
\midrule
  & \multicolumn{3}{c}{\textbf{prs\_Arab}} & \multicolumn{3}{c}{\textbf{tat\_Cyrl}} & \multicolumn{3}{c}{\textbf{mag\_Deva}} & \multicolumn{3}{c}{\textbf{lus\_Latn}} \\
\cmidrule{2-13}
  & Low  & Mid & High & Low  & Mid & High & Low  & Mid & High & Low  & Mid & High \\
\hline
m2m\_100      & -        & -       & -        & -        & -       & -        & -        & -       & -        & -        & -        & -       \\
Finetune      & 14.44 & 16.34 & 18.08 & 13.41 & 14.68 & 18.86 & 14.65 & 16.49 & 20.81 & 16.46 & 18.24 & 21.29 \\
Extend\_Vocab & 15.94 & 19.51 & 20.34 & 14.74 & 16.04 & 20.96 & 15.28 & 18.95 & 22.54 & 18.29 & 19.49 & 22.77 \\
KD            & 16.82 & 20.58 & 21.29 & 16.16 & 18.61 & 21.83 & 16.93 & 20.72 & 23.48 & 17.09 & 20.69 & 23.45 \\
Adapter       & 18.29 & 24.25 & 21.69 & 17.96 & 18.44 & 22.63 & 17.80 & 21.19 & 23.37 & 18.15 & 21.44 & 24.60 \\
MMA           & 19.70 & 24.52 & 22.35 & 19.72 & 20.09 & 23.60 & 19.69 & 21.76 & 24.37 & 19.38 & 22.40 & 25.56 \\
\midrule
\md           & \textbf{26.96} & \textbf{29.71} & \textbf{33.28} & \textbf{26.69} & \textbf{29.33} & \textbf{32.27} & \textbf{27.01} & \textbf{29.50} & \textbf{31.92} & \textbf{29.55} & \textbf{32.42} & \textbf{34.99} \\
\bottomrule
\end{tabular}
\caption{\label{tab:q1_chrf}
\textbf{Main Results (the answer of Q1)}: Average chrF++ scores for different categories in two directions on  FLORES-200 benchmark. The original languages indicates the languages already supported in m2m\_100. 
$^\ddag$ denotes significant over original \textit{m2m\_100} model at 0.05/0.01, evaluated by bootstrap resampling~\cite{koehn-2004-statistical}.
}
\vspace{-1.2em}
\end{table*}
\begin{table*}[h]
\small
\centering
\begin{tabular}{l|ccccccccc}
\toprule
\multicolumn{10}{c}{\textbf{Extended model trained from prs\_Arab to original languages}} \\
\midrule
              & L2L  & L2M  & L2H  & M2L   & M2M   & M2H   & H2L  & H2M   & H2H   \\
\midrule
m2m\_100      & \textbf{18.03} & \textbf{22.69} & \underline{30.82} & \underline{34.40} & \underline{34.37} & \textbf{35.45} & \underline{31.66} & \underline{33.12} & \textbf{36.16} \\
Finetune      & 13.45 & 11.64 & 15.64 & 17.78 & 22.95 & 25.57 & 23.10 & 22.04 & 27.40 \\
Extend\_Vocab & 7.35  & 16.09 & 13.16 & 15.49 & 20.89 & 20.58 & 17.79 & 19.02 & 21.06 \\
KD            & 13.25 & 15.70 & 25.85 & 30.24 & 31.33 & 31.88 & 27.02 & 27.37 & 31.69 \\
Our           & \underline{17.58} & \underline{22.17} & \textbf{31.06} & \textbf{34.78} & \textbf{34.61} & \underline{35.35} & \textbf{31.71} & \textbf{33.15} & \underline{36.01} \\
\midrule\midrule
\multicolumn{10}{c}{\textbf{Extended model trained from original languages to mag\_Deva}} \\
\midrule
              & L2L  & L2M  & L2H  & M2L   & M2M   & M2H   & H2L  & H2M   & H2H   \\
\midrule
m2m\_100      & \textbf{16.77} & \textbf{19.95} & \textbf{28.52} & \underline{33.09} & \underline{33.43} & \textbf{37.24} & \underline{28.27} & \textbf{34.95} & \underline{36.38} \\
Finetune      & 8.19  & 12.18 & 14.56 & 18.96 & 23.96 & 22.69 & 19.00 & 24.08 & 23.62 \\
Extend\_Vocab & 8.51  & 11.93 & 9.19  & 18.53 & 20.46 & 20.25 & 15.83 & 21.80 & 19.80 \\
KD            & 13.17 & 15.95 & 21.58 & 26.68 & 28.03 & 31.07 & 21.27 & 29.73 & 32.03 \\
Our           & \underline{16.12} & \underline{18.26} & \underline{28.09} & \textbf{33.22} & \textbf{33.94} & \underline{37.05} & \textbf{28.34} & \underline{34.87} & \textbf{36.42} \\
\bottomrule
\end{tabular}
\caption{
\label{tab:q2_chrf}
\textbf{Main Results (the answer of Q2):} Average BLEU scores of the extended model on 9 original language pairs grouped by available resources on both source and target sizes (\textbf{L}ow, \textbf{M}id, \textbf{H}igh). $\Delta$ indicates the difference between \textit{m2m\_100} and our approach. \textbf{Bold} and \underline{underlined} numbers indicates the best and second-best results respectively. We do not include the results of \textit{Adapter} and \textit{MMA} method, because they fix the parameters of original model and the results are the same as in original m2m\_100.
}
\end{table*}

\noindent \textbf{Q1: Successfully Extending to a New Language}

\textbf{Training Directions.}
Comparing the two translation directions in Tables~\ref{tab:q1} and \ref{tab:q1_chrf}, we observe that translations from the new language to the original languages perform better than the reverse. 
This is likely due to weaker decoding performance in the new language. \citet{he2019hard} noted that machine translation decoders are more sensitive to noisy inputs than encoders.
As the new language has less data available, training an efficient decoder from scratch is challenging, introducing additional noise into the translation process and impairing performance.
Table~\ref{tab:copy_off} further illustrates this issue. However, translation from the new language into the original languages does not suffer from these challenges, as the new language shares some vocabulary with the original languages on the source side.

\begin{figure}[!h]
\centering
        \includegraphics[width=\linewidth]{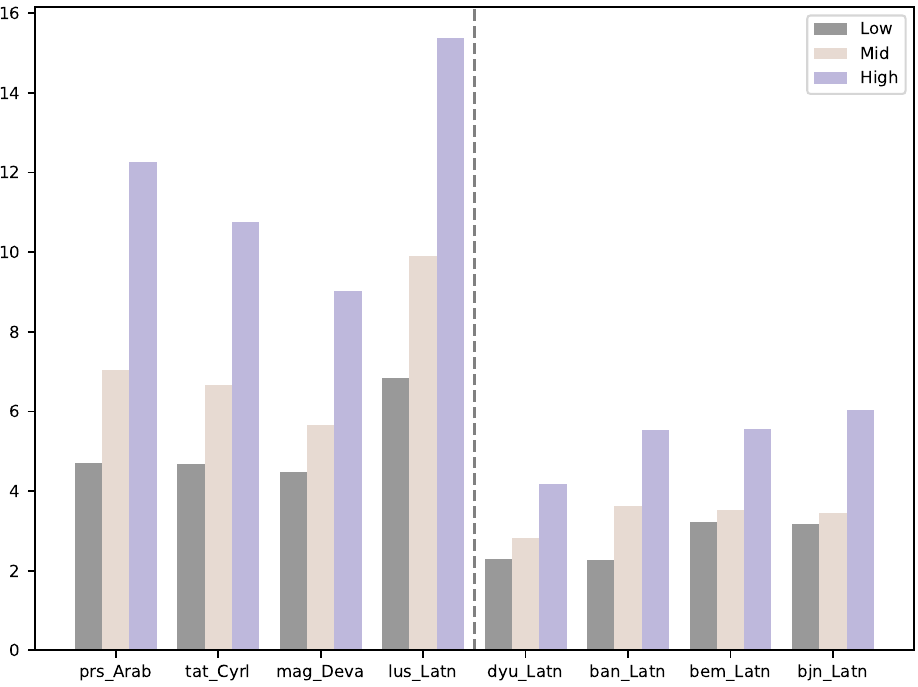}
	\caption{
 \textbf{Corpus size analysis:} the four languages on the left are those used in the main experiment with corpus sizes in the tens of millions, while the latter four are comparison languages with corpus sizes in the hundreds of thousands.
 }
    \label{fig:corpus_size}
 \vspace{-1em}
\end{figure}

\textbf{Corpora Sizes.}
In addition to the four primary languages tested (Dari, Tatar, Magahi, and Mizo), we also evaluate four languages with smaller corpora (Dyula, Balinese, Bemba, and Banjar). 
Corpus statistics are shown in Table~\ref{tab:app_data_sta}.
As demonstrated in Figure~\ref{fig:corpus_size}, languages with smaller corpora perform worse, while languages with larger datasets, like those in the main experiments, achieve robust performance under our imitation learning framework.
This is because larger corpora allow the model to reach a sufficient level of proficiency, whereas small corpora can cause premature convergence, resulting in suboptimal performance.

\textbf{Language Categories.}
Our method performs better with language pairs where the original language is high-resource.
This is due to the superior performance of the original m2m\_100 model in translating high-resource languages.
In contrast, translations involving low-resource languages produce BLEU scores mostly below 5, and the noisy nature of such data further degrades the quality of translations when using low-resource languages.

\noindent \textbf{Q2: avoiding catastrophic forgetting}

\textbf{Baselines.}
Tables~\ref{tab:q2} and \ref{tab:q2_chrf} indicate that all baseline methods suffer from catastrophic forgetting, where the focus on the new language deteriorates performance on the original language pairs.

\textbf{Training Directions.}
Comparison of training directions in Tables~\ref{tab:q2} and \ref{tab:q2_chrf} shows that extending the source side produces better results than extending the target side.
For instance, the extended model trained from prs\_Arab performs better than that trained from original languages to mag\_Deva.
This phenomenon is similar to the conclusion drawn in Table~\ref{tab:q1}, i.e., the performance of the extended model trained from the new language to the original language surpasses that of the model trained in the opposite direction, reinforcing the consistency of the findings.

\textbf{Language Categories.}
Language pairs involving high-resource target languages (e.g., \textit{L2H}, \textit{M2H}, \textit{H2H}) consistently outperform other six translation categories, especially in \textit{H2H} pairs due to abundant training data and higher weights assigned during imitation learning.

\begin{figure*}[!h]
\centering
        \includegraphics[width=\linewidth]{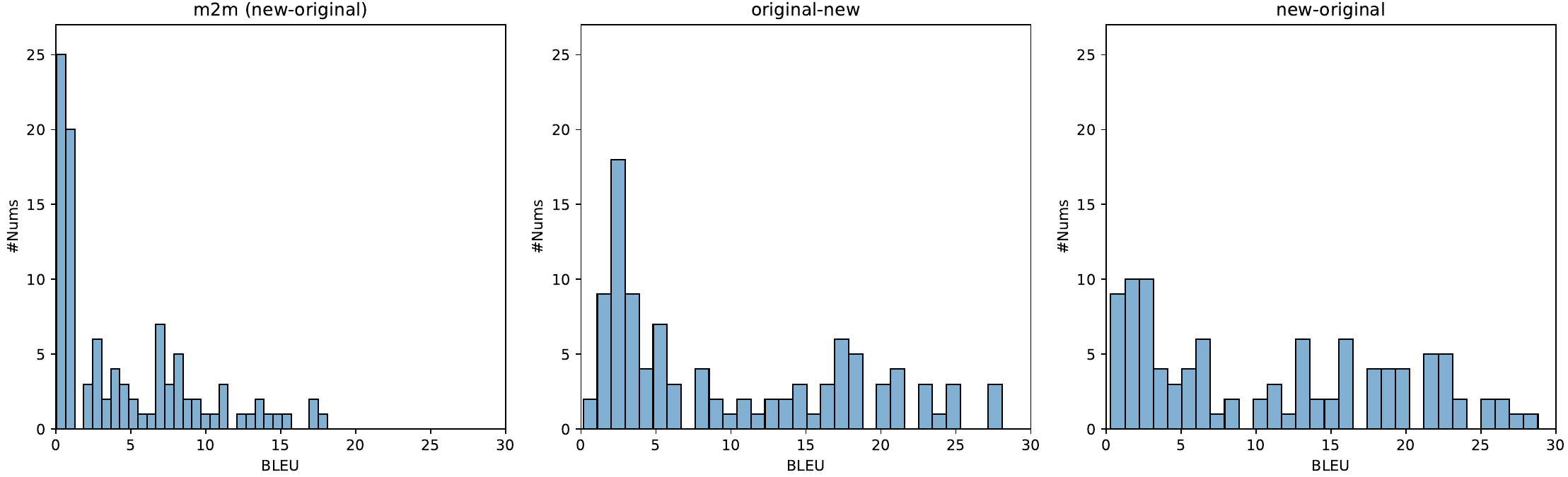}
	\caption{
\textbf{BLEU score distribution statistics:} the original model, the extended model for translations from the original languages to the new language, and the extended model for translations from the new language to the original languages.
 }
    \label{fig:bleu_dis}
 \vspace{-1em}
\end{figure*}

\section{BLEU Distribution}
\label{appendix:bleu_dis}
In Figure 3, we present the BLEU score distributions for the original m2m\_100 model, translations from the original languages to the new language, and from the new language to the original languages. The figure shows that our extension method reduces the distribution of BLEU scores in the 0-2 range.

\section{On-the-Fly Approach}
\label{appendix:on_the_fly}
After constructing the pseudo $k$-way parallel data (Section~\ref{sec:data_construct}), an intuitive idea is to use this data to update the parameters of the expert MNMT model.
This is known as the \textit{On-the-Fly} approach, which involves using the same model to construct the pseudo-corpus and updating the parameters.
However, this approach faces the following challenges:
i) The pseudo corpus introduces noise that has a significant impact on the training process, particularly when dealing with low-resource languages.
Related experiments can be found in Table~\ref{sec:ablation}.
ii) Similarly, the introduction of noisy data directly affects the representation of the selected $k$ original languages in the MNMT model, leading to a substantial impact on the performance of the original language pairs.
Our results regarding this aspect can be found in Table~\ref{tab:ablation_lw_on}.
To mitigate these challenges, we treat the original MNMT model as an expert and keep it frozen, while we train a separate learner model with the ultimate objective of acquiring the capability to translate between the new language and the original languages\footnote{We consider two directions: either train the extended model from the new to the original languages or train the extended model from the original languages to the new language.} by weighting the language (Section~\ref{sec:data_dis}) and mimicking translation behavior (Section~\ref{sec:imit}) of the expert model.

\begin{figure}[!h]
\centering
        \includegraphics[width=\linewidth]{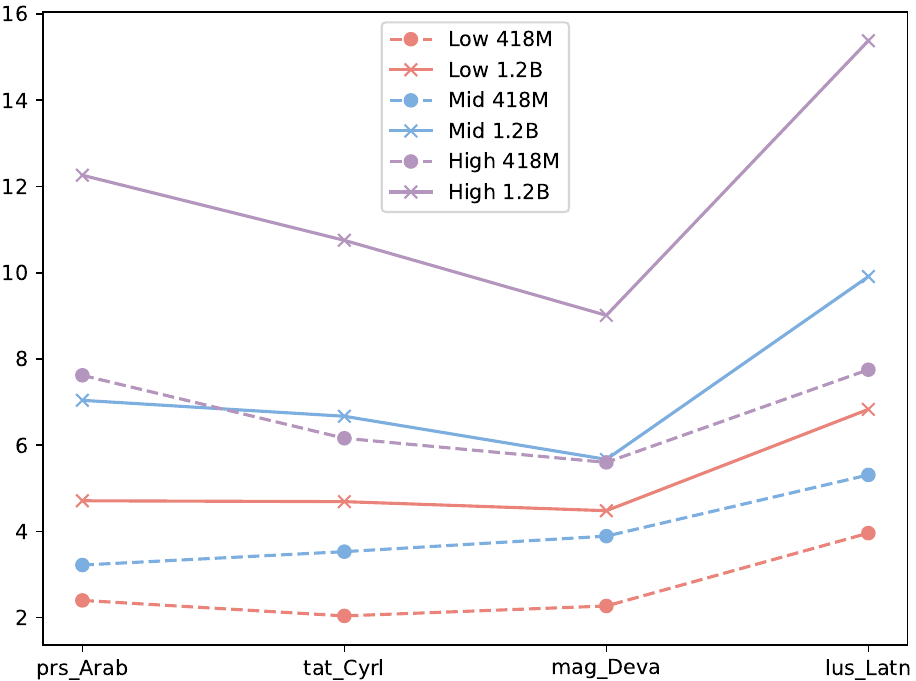}
	\caption{
\textbf{Model size analysis:} a comparison of the performance between the 418M model and the 1.2B model (used in main experiments) across three language categories.
 }
    \label{fig:model_size}
 \vspace{-1em}
\end{figure}

\begin{figure*}[!th]
\vspace{-1em}
\centering
        \includegraphics[width=\linewidth]{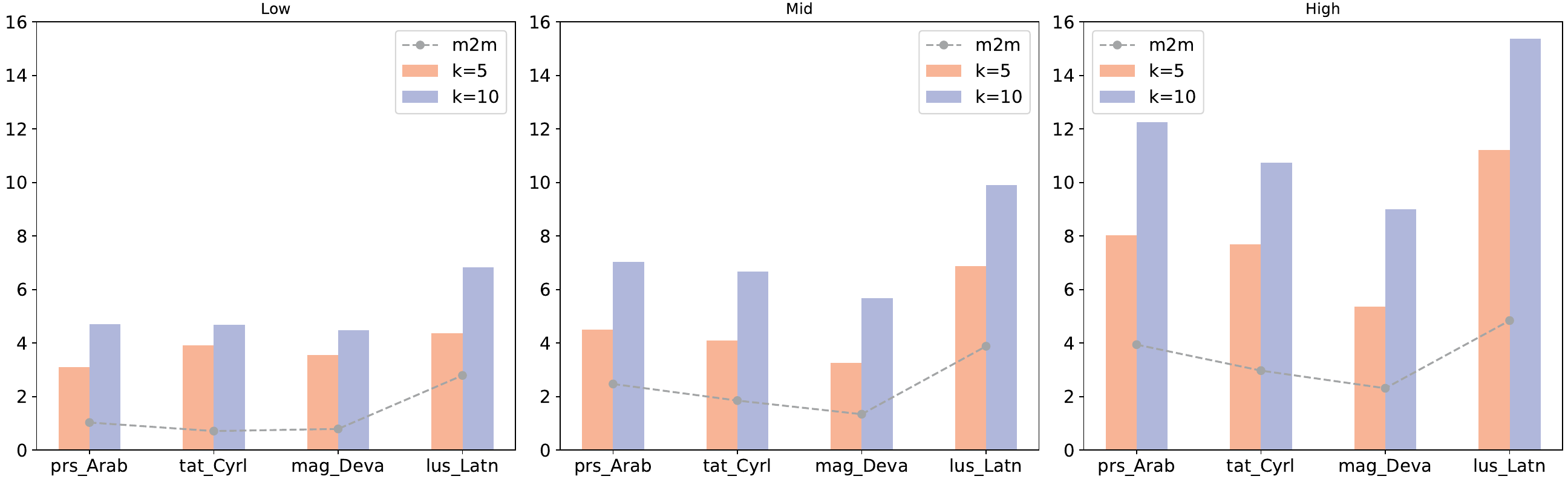}
	\caption{
\textbf{k-value analysis:} performance comparison between k values of 5 and 10.
 }
    \label{fig:k_values}
 \vspace{-1em}
\end{figure*}

\section{Copy and Off-Target Problem}
\label{appendix:cr_otr}
In contrast to \cite{liu-etal-2021-copying}, we consider two distinct types of copying behaviors:
i) the proportion of tokens copied from the source sentence;
ii) the ratio of consecutively repeated words in the generated target sentences.
The total copying ratio (CR) can be formulated as follows:
\begin{align}
CR=\frac{\sum_{i=1}^{T} cs(i)}{\sum_{i=1}^{T} count(i)} + \frac{\sum_{i=1}^{T} rt(i)}{\sum_{i=1}^{T} count(i)}
\end{align}
where $cs(\cdot)$ is number of tokens \underline{c}opied from the \underline{s}ource sentence ($i$), $rt(\cdot)$ is the number of consecutive \underline{r}epeated tokens in the generated \underline{t}arget sentences and $count(\cdot)$ is the number of
tokens in the generated target sentence. $T$ is the number of sentences in the test set.

To quantify the extent of off-target behaviors, we compute the ratio of off-target sentences in the translation outputs using the following formula:
\begin{align}
OTR=\frac{\sum_{i=1}^{T} ot(i)}{T}
\end{align}
where $ot(\cdot)$ is a function that judges whether a sentence belongs to an incorrect language\footnote{We use language identification from NLLB: \url{https://dl.fbaipublicfiles.com/nllb/lid/lid218e.bin}}.

\begin{table}[!h]
\small
\centering
\resizebox{\columnwidth}{!}{
\begin{tabular}{l|ccc|ccc}
\toprule
& \multicolumn{3}{c}{\textbf{eng-rus}} & \multicolumn{3}{c}{\textbf{eng-wol}} \\
\cmidrule(lr){2-4}\cmidrule(lr){5-7}
              & chat  & health & news  & chat & health & news \\
\midrule
m2m\_100      & 26.14 & 35.28  & 26.17 & 4.26 & 5.35   & 4.59 \\
Finetune      & 22.46 & 30.59  & 22.83 & 2.45 & 4.01   & 2.45 \\
Extend\_Vocab & 21.55 & 29.06  & 22.04 & 2.20 & 3.88   & 2.21 \\
KD            & 22.98 & 30.62  & 23.04 & 2.74 & 4.86   & 2.85 \\
Adapter       & 21.76 & 29.74  & 22.82 & 2.86 & 3.47   & 1.77 \\
MMA           & 21.63 & 29.26  & 22.40 & 2.22 & 3.17   & 2.00 \\
\midrule
Our           & \textbf{27.81} & \textbf{36.17}  & \textbf{27.31} & \textbf{4.74} & \textbf{6.22} & \textbf{4.78} \\
\bottomrule
\end{tabular}}
\caption{
\label{tab:domain_trans}
\textbf{Domain Transfer}: evaluate the zero-shot domain transfer on the extended model for the \textit{prs\_Arab} language.
}
\end{table}

\section{Zero-Shot Domain Transfer}
\label{appendix:domain}
To explore the zero-shot domain transfer~\cite{lai-etal-2022-4,lai-etal-2022-improving-domain} capacity of our models, we utilize the extended model for the \texttt{prs\_Arab} language in different domains.
All experiments are conducted on the FLORES-200 multi-domain dataset. The corresponding results can be found in Table~\ref{tab:domain_trans}.
We obeserve that our approach demonstrates strong domain transfer capabilities, surpassing the baseline systems, even when applied to the original language pairs such as \textit{eng-rus} and \textit{eng-wol}.

\section{Model Size Analysis}
\label{appendix:model_size}
To evaluate the performance of our approach across different model sizes, we compared the \texttt{418M} and \texttt{1.2B} versions of the m2m\_100 model.
As shown in Figure~\ref{fig:model_size}, the performance of the 1.2B model is significantly better than that of the 418M model.
For instance, in low-resource languages, the average BLEU score for the 418M model ranges between 2 and 3, whereas the 1.2B model achieves an average BLEU score of 4 to 7.
This is particularly meaningful because models with very low BLEU scores tend to generate random text, where a BLEU score of 0.5 and 1.3 often do not represent a substantial difference.
Previous work by~\citet{zhang-etal-2020-improving} and~\citet{lin-etal-2021-learning} has confirmed that BLEU can struggle to accurately reflect translation quality when scores are very low.
The 1.2B model used in our main experiments effectively addresses the issue of low BLEU scores, as evidenced by the distribution of BLEU scores. 
This improvement is primarily due to the overall superior performance of the 1.2B model compared to the 418M model, further reinforcing the advantages of our imitation learning framework.

\section{Impact of $k$}
\label{appendix:k_value}
To verify the impact of the value of k on our approach, as discussed in Section~\ref{sec:methods}, we conduct experiments with k=5 and k=10.
As shown in Figure~\ref{fig:k_values}, the performance is better when k=10 compared to k=5.
This is primarily because a larger k allows more language pairs to be imitated in each batch, resulting in better overall performance.

\end{document}